\documentclass{article}

\usepackage[preprint]{neurips_2024}

\usepackage[utf8]{inputenc} 
\usepackage[T1]{fontenc}    
\usepackage{url}            
\usepackage{booktabs}       
\usepackage{amsfonts}       
\usepackage{nicefrac}       
\usepackage{microtype}      
\usepackage{xcolor}         

\newcommand{\ignore}[1]{}
\RequirePackage{amsthm,amsmath,amsfonts,amssymb}
\RequirePackage{natbib}
\RequirePackage[colorlinks,citecolor=blue,urlcolor=blue]{hyperref}
\usepackage{algorithm}
\usepackage{algorithmic}
\RequirePackage{graphicx,xcolor}
\usepackage{siunitx}
\usepackage{capt-of}
\usepackage{booktabs}
\usepackage{varwidth}
\usepackage{enumitem}   
\usepackage{comment}
\usepackage{lineno}
\usepackage{amsbsy}
\usepackage{blindtext}

\numberwithin{equation}{section}
\theoremstyle{plain}

\newtheorem{theorem}{Theorem}[section]
\newtheorem{proposition}{Proposition}[section]
\newtheorem{lemma}[theorem]{Lemma}

\newtheorem{definition}[theorem]{Definition}

\theoremstyle{remark}

\newtheorem{remark}{Remark}[section]

\RequirePackage[utf8]{inputenc} 
\RequirePackage[T1]{fontenc}    
\RequirePackage{multicol,booktabs,longtable, multirow, makecell}       
\RequirePackage{bm}  
\RequirePackage{nicefrac}       
\usepackage{textcomp}
\RequirePackage{microtype}      
\RequirePackage{cleveref}       
\RequirePackage{psfrag,epsf}
\RequirePackage{caption,subcaption}
\usepackage{enumitem}

\def\R{\mathbb{R}}

\newcommand{\bad}{\text{bad}}

\newcommand{\dN}{\mathcal{N}}
\newcommand{\dunif}{\mathcal{U}}

\newcommand{\wh}{\widehat}
\newcommand{\wt}{\widetilde}

\newcommand{\IP}{\mathbb{P}}

\newcommand{\fancyname}{\texttt{NoisyHead}}

\definecolor{mygreen}{rgb}{0.09,0.62,0.45} 
\definecolor{myorange}{rgb}{0.83,0.37,0.10} 
\definecolor{myblue}{rgb}{0.06,0.45,0.69}  
\definecolor{myred}{rgb}{0.94,0.20,0.13}
\definecolor{mypurple}{RGB}{76,0,153} 

\title{How Private is Your Attention? \\ Bridging Privacy with In-Context Learning}

\author{%
  Soham Bonnerjee\\
  \texttt{sohambonnerjee@uchicago.edu} \\
  \And
  Yeon Zhen Wei (Kingsley) \\
  \texttt{yeon@uchicago.edu} \\
  \And
  Anna Asch\\
  \texttt{aasch@uchicago.edu} \\
  \And
  Sagnik Nandy \\
  \texttt{sagnik@uchicago.edu} \\
  \And
  Promit Ghosal \\
  \texttt{promit@uchicago.edu} \\
}

\begin{document}

\maketitle

\begin{abstract}
In-context learning (ICL)—the ability of transformer-based models to perform new tasks from examples provided at inference time—has emerged as a hallmark of modern language models. While recent works have investigated the mechanisms underlying ICL, its feasibility under formal privacy constraints remains largely unexplored. In this paper, we propose a differentially private pretraining algorithm for linear attention heads and present the first theoretical analysis of the privacy–accuracy trade-off for ICL in linear regression. Our results characterize the fundamental tension between optimization and privacy-induced noise, formally capturing behaviors observed in private training via iterative methods. Additionally, we show that our method is robust to adversarial perturbations of training prompts, unlike standard ridge regression. All theoretical findings are supported by extensive simulations across diverse settings.
\end{abstract}

\section{Introduction}
Attention-based models, particularly large language models (LLMs), have demonstrated remarkable capabilities in performing \emph{in-context learning}~\citep{brown2020language, lieber2021jurassic, rae2021scaling, black2022gptneox, bubeck2023sparks}. This paradigm has transformed human-AI interaction, enabling AI models to tackle complex tasks without explicit parameter updates. A growing body of theoretical work has aimed to explain this emergent behavior~\citep{dong2022general, akyurek2022learning, garg2022what, wang2023incontext,xie2022explanation}, often using simplified settings. These studies suggest that transformers can implicitly infer patterns or rules from training examples in the prompt and apply them to new, related inputs during inference.

The growing use of LLM-based agents in sensitive domains such as medicine~\citep{li2025privacy, dennstadt2025implementing} and psychology~\citep{ke2024exploring} underscores the urgent need for robust privacy safeguards. In particular, model providers must prevent adversaries from extracting sensitive training data, a risk highlighted by recent work demonstrating that LLMs can memorize and reveal specific examples when prompted adversarially~\citep{carlini2021extracting, carlini2021membership, tirumala2022memorization}. A principled approach to mitigating such leakage is \emph{differential privacy} (DP)~\citep{dwork2006calibrating}, which ensures that an algorithm’s output remains nearly unchanged when a single training point is modified. This is typically achieved by injecting calibrated noise to limit individual influence.

However, integrating privacy-preserving mechanisms into the pretraining process of a transformer inevitably degrades the downstream performance of in-context learning on test prompts. This trade-off motivates a rigorous study of the {cost of privacy} of \textit{in-context differentially-private} algorithms: what additional error is incurred at test time?

\subsection{Main Results}
We study the effect of differentially-private pretraining on in-context learning (ICL) for linear regression, where each data point is a noisy linear response to input features. We propose a differentially-private pretraining algorithm for a linear attention head that performs ICL—predicting the response for a query input by attending to a sequence of labeled input-output examples. The model is trained on $N$ prompts, each containing $L$ feature-response pairs sampled from a noisy linear model, and optimized to minimize squared prediction error on the query token. To enforce privacy, we apply the Gaussian mechanism—gradient clipping followed by additive noise—commonly used in private empirical risk minimization~\citep{dwork2006calibrating, chaudhuri2011differentially, abadi2016deep, tcai-cost-of-privacy}. Our method, \fancyname{} (Algorithm~\ref{algo:DP2}, Section~\ref{pre-train}), formalizes this approach.

We define the \emph{cost of privacy} as the difference, between attention heads trained with and without privacy constraints, in average prediction error of the response to a query token from a held-out test prompt. Our main theoretical result characterizes how the \emph{cost of privacy} scales with the number of training prompts $N$, the prompt length $L$, the token dimension $D$, and the privacy parameters $(\varepsilon, \delta)$. We state it informally below:

\begin{theorem}[Informal]
In the low dimensional regime, when $L$ and $\sqrt{N}$ are asymptotically of same order and $D = O(1)$, the cost of privacy satisfies
\[
\text{Cost of Privacy} \lesssim \frac{1}{N^{3/2}L^2}\frac{\log(1/\delta)}{\varepsilon^2}.
\]
In the high dimensional regime, when $N/D^2 = O(1)$ and $L / D = O(1)$, the cost of privacy scales as
\[
\text{Cost of Privacy} \lesssim \frac{D^2}{N^2L^2}\frac{\log(1/\delta)}{\varepsilon^2},
\]
up to $\mathrm{polylog}$ factors.
\end{theorem}
A formal version of this result is presented in Theorem~\ref{cor:dic_behav}, followed by a detailed discussion of its implications. The theorem highlights that the cost of privacy exhibits fundamentally different behavior in the low- and high-dimensional regimes. In the \emph{low-dimensional} setting, the minimax cost of privacy for learning a linear model from $L$ labeled data points is known to scale as
$
(\varepsilon L)^{-2} \cdot \log(1/\delta),
$
as established in~\cite{tcai-cost-of-privacy}. The result above shows that leveraging contextual data reduces this cost to
$
N^{-3/2}(\varepsilon L)^{-2} \cdot \log(1/\delta).
$
However, because test-time prediction requires learning an unseen coefficient vector $w$, we do not achieve the rate
$
N^{-2}(\varepsilon L)^{-2} \cdot \log(1/\delta),
$
which would be expected if the coefficient was identical across all training and test prompts. In contrast, in the \emph{high-dimensional} regime, where the feature dimension scales with the number of prompts $N$, we incur an additional multiplicative factor of $\sqrt{N}$ in the denominator due to the increased complexity of the learning problem.


We also show that our private pretraining procedure is more robust to adversarial perturbations of training prompts than its non-private counterpart. When a fraction of prompts are corrupted, the prediction risk on test instances remains significantly more stable under our method — a property especially relevant given recent concerns about adversarial attacks in LLMs~\citep{anwar2024adversarial}.

Our key contributions are as follows:
    
    \noindent $\mathbf{(1)}$ We propose a differentially-private pretraining algorithm (\fancyname) based on the \emph{Gaussian mechanism} for training linear attention heads to perform in-context learning in linear regression (see Algorithm \ref{algo:DP2}). Our method is motivated by the differentially-private stochastic gradient descent algorithm~\citep{abadi2016deep}, containing a tuned noise-injection at the gradient steps.
    
    \noindent $\mathbf{(2)}$ We provide a detailed theoretical analysis of the excess risk incurred by enforcing differential privacy during pretraining in Theorem~\ref{thm:main_thm}. In particular, it characterizes the privacy–utility trade-off, quantifying the impact of privacy constraints on the prediction error of \fancyname\ across any number of iterations $T$ of the algorithm. This trade-off exhibits dichotomous behavior depending on how the feature dimension $D$ scales with the number of training samples $N$. We identify two distinct regimes: one where $D = O(\log N)$ and another where $N/D^2 = O(1)$. These lead to qualitatively different error decay rates with respect to $N$, $L$, and $D$, as formalized in Theorem~\ref{cor:dic_behav}. In the over-parametrized setting when $N, L^2, D^2$ are asymptotically of the same order, we show that there is a delicate interplay between the number of training iterations and the generalization error on unseen prompts. Due to the injection of noise at each iteration, longer training can degrade generalization, necessitating careful selection of the number of optimization steps. This highlights the importance of ``early stopping'' for the algorithm. See Proposition~\ref{prop1} and the following remark for related discussion.

    
    \noindent $\mathbf{(4)}$ We establish that \fancyname{} exhibits a notable robustness property under adversarial perturbations to the training data, particularly during the pretraining stage. Compared to the baseline method proposed in~\cite{lu2024context}, our approach shows significantly less degradation in generalization error in the presence of such perturbations. In the baseline setting, where the linear attention module is pretrained using ridge regression, even moderately large perturbations can induce a distributional shift in the training data, leading to inaccurate estimation of model weights and consequently poor generalization. In contrast, \fancyname{} incorporates a truncation mechanism that clips responses, predictors, and weights within prescribed compact sets. This simple yet effective step restricts the influence of corrupted or outlying data points, enhancing robustness to adversarial noise introduced during training. Theoretical support for this robustness is provided in Theorem~\ref{thm:robustness-thm}.

    \noindent $\mathbf{(5)}$ We conduct a comprehensive empirical study to validate the theoretical predictions of our analysis. In both low- and high-dimensional regimes (Section~\ref{se:N vs eps}), we demonstrate that the excess prediction risk of \fancyname\ decays with increasing sample size and privacy parameter, consistent with the rates derived in Theorem~\ref{cor:dic_behav}. Moreover, in the overparameterized regime (Section~\ref{se:over-parametrized}), our experiments reveal a distinct phase transition in the generalization error: initially decreasing due to optimization, but eventually increasing due to cumulative noise from differential privacy. This phenomenon, visualized in Figure~\ref{fig:phase-transition}, substantiates the theoretical trade-off outlined in Proposition~\ref{prop1} and underscores the critical role of early stopping. Finally, robustness experiments (Section~\ref{se:robustness-simu}) confirm that \fancyname\ maintains stable performance under adversarial perturbations, while ridge-based pretraining degrades significantly. These results highlight the practical utility of our method and affirm the relevance of our theoretical contributions in realistic settings.


\subsection{Related literature and notations}

Since its introduction by \citet{dwork2006calibrating}, differential privacy has become a cornerstone of privacy-preserving machine learning, inspiring a wide range of algorithms across classical and deep learning tasks~\citep{tcai-cost-of-privacy, wang2019sparse, gu2024differential, jain2013differentially, ni2016detecting, ji2019differentially, abadi2016deep, feldman2018privacy}. In parallel, recent work has explored the in-context learning (ICL) capabilities of transformers, demonstrating that pretraining enables them to emulate diverse algorithms—including ridge regression, generalized linear models, Lasso, and neural networks—purely from contextual examples~\citep{dai2023transformers}, with theoretical insights provided for linear attention models by \citet{bartlett-icl-lm-24} and \citet{lu2024context}. Despite significant advances in both areas, their intersection remains underexplored: while prior work has investigated differentially-private pretraining for transformers~\citep{majmudar2022differentially, yu2023dptraining, li2021large} and evaluated the privacy properties of language models~\citep{hoory2021learning, anil2021large}, the impact of privacy on downstream ICL performance has not been theoretically analyzed. This paper bridges this gap by providing the first rigorous analysis of how imposing differential privacy during pretraining influences the in-context learning capabilities of attention-based models.

\subsubsection{Notation} In this paper, we denote the set $\{1, \ldots, n\}$ by $[n]$. $d$-dimensional Euclidean space is $\R^d$, with $\R^d_{>0}$ the positive orthant. The set of $m \times n$ real matrices is $\R^{m \times n}$, and $\mathbb{S}^{d-1}$ denotes the $d$-dimensional unit sphere. The Frobenius norm of a matrix $A$ is $\|A\|_F$, and $\langle \cdot, \cdot \rangle$ denotes the standard inner product. We write $a_n \lesssim b_n$ if $a_n \le C b_n$ for some constant $C > 0$, and $a_n \asymp b_n$ if $C_1 b_n \le a_n \le C_2 b_n$ for some constants $C_1, C_2 > 0$. We also write $a_n \asymp b_n$ as $a_n = \Theta(b_n)$.

\section{Problem Formulation}
We consider a set-up where we observe a sequence of labeled tokens $\{(y_i,x_i): i \in \{1,\ldots,L\}\}$, for $x_i \overset{i.i.d}{\sim} \dunif(\mathbb S^{D-1})$ and $y_i = w^\top x_i + \epsilon_i$, with $w \sim \dN_D(0,\mathbb{I}_D)$ and $\epsilon_i \overset{i.i.d}{\sim} \dN(0,\tau^2)$. Here $\dunif(\mathbb S^{D-1})$ denotes the uniform distribution on the $D$-dimensional hypersphere and $\dN_k(\mu,\Sigma)$ denotes the $k$ dimensional normal distribution with mean $\mu$ and covariance $\Sigma$. For a test token $(y_{L+1},x_{L+1})$ generated independently from the same distribution as the training tokens, we want to predict $y_{L+1}$ based on $x_{L+1}$. 

This setting was used by \citet{bartlett-icl-lm-24} and \citet{lu2024context}, both of whom considered the noiseless case of $\tau^2=0$. As proposed therein, we embed the prompt as
\begin{align}
\label{eq:def_prompt}
E = \begin{pmatrix}
    x_1 & x_2 & \cdots & x_L & x_{L+1} \\
    y_1 & y_2 & \cdots & y_L & 0
\end{pmatrix} \in \mathbb{R}^{(D+1) \times (L+1)}.
\end{align}
This matrix is passed through a single linear attention head as follows:
\begin{align}\label{eq:lsa}
    f(E; \theta) = E + W^{PV} E \cdot \frac{E^\top W^{KQ} E}{L},
\end{align}
where $\theta = (W_{PV}, W_{KQ})$ with $W_{PV} \in \R^{(D+1) \times (D+1)}$ and $W_{KQ} \in \R^{(D+1) \times (D+1)}$.
The prediction of the query response is given by the $(D+1, L+1)$-th entry of $f(E; \theta)$; that is,
$\wh y_{L+1}(E) = (f(E; \theta))_{(D+1, L+1)}$.
We aim to learn the parameters of the model $f(E;\theta)$ by pretraining the model based on $N$ training prompts $\left\{(y_{k,1},x_{k,1}),\ldots,(y_{k,L},x_{k,L}),(y_{k,L+1},x_{k,L+1})\right\}_{k=1}^N$,
where the $L+1$-th token is the query token. Putting the prompts into matrices $E_1,\ldots,E_N,$ we have
\begin{align*}
    E_k:= \begin{pmatrix}
        x_{k,1} & x_{k,2} & \cdots & x_{k,L} & x_{k,L+1}\\
        y_{k,1} & y_{k,2} & \cdots & y_{k,L} & 0
    \end{pmatrix} \in \mathbb{R}^{(D+1) \times (L+1)}.
\end{align*}
Now we minimize the standard loss function $\mathcal L(\theta)=\frac{1}{2N}\sum_{i=1}^{N}(\wh y_{L+1}(E_k)-y_{k,L+1})^2.$
The predictor $(f(E; \theta))_{(D+1, L+1)}$ can be simplified by linear algebra to
\begin{align}\label{eq:y_hat}
    \hat{y}_{L+1}:= [(f(E;\theta)_{[D+1, L+1]}]=\left[\left(w_{21}^{P V}\right)^{\top} \quad w_{22}^{P V}\right] \left(\frac{E E^{\top}}{L}\right)\begin{bmatrix}W_{11}^{K Q} \\
    \left(w_{21}^{K Q}\right)^{\top}\end{bmatrix} x_{L+1},
\end{align}
where we have used the matrices $W^{PV}$ and $W^{KQ}$, partitioned as follows:
\[ W^{P V}=\begin{bmatrix}
W_{11}^{P V} & w_{12}^{P V} \\
\left(w_{12}^{P V}\right)^{\top} & w_{22}^{P V}
\end{bmatrix}, \quad W^{K Q}=\begin{bmatrix}
W_{11}^{K Q} & w_{12}^{K Q} \\
\left(w_{12}^{K Q}\right)^{\top} & w_{22}^{K Q}
\end{bmatrix},\]
with $W_{11}^{PV}, W_{11}^{KQ}\in \R^{D\times D}$, $w_{21}^{PV}, w_{21}^{KQ}\in \R^{D}$, and $w_{22}^{PV}, w_{22}^{KQ}\in \R$. 
The quadratic form \eqref{eq:y_hat} can be expanded to yield
\begin{align} \label{eq:yhat_old}
&\hat{y}_{L+1} = \frac{1}{L} \langle x_{L+1}, Q^{(1)}_W + Q^{(2)}_W \rangle, 
\end{align}
where $Q^{(1)}_W := w^{PV}_{22} W^{KQ}_{11} \sum_{i=1}^L y_i x_i + w^{PV}_{22} w^{KQ}_{12}\sum_{i=1}^L y_i^2 $ and $Q^{(2)}_W:=W^{KQ}_{11} \sum_{i=1}^{\ell+1} x_i x_i^\top w^{PV}_{12} + w^{KQ}_{12} \sum_{i=1}^\ell y_i x_i^\top w^{PV}_{12}$. Following~\cite{yu2023dptraining} and~\cite{bartlett-icl-lm-24}, we adopt the assumption that $w^{KQ}_{12}=0$ and $w^{PV}_{12} = 0$ throughout this paper. This particular choice is also explained in Section \ref{se:choice}. Let us define
\begin{align}
\label{eq:construct_meta_varb}
\Gamma = w_{22}^{PV} W_{11}^{KQ} \in \mathbb{R}^{D \times D}, \quad \text{and} \quad Z = \frac{1}{L} x_{L+1} \sum_{i=1}^L y_i x_i^\top \in \mathbb{R}^{D \times D}.
\end{align}
With this definition of $\Gamma$ and $Z$, the predictor $\widehat{y}$ simplifies to the inner product $\widehat{y} = \langle \Gamma, Z \rangle,$ and we train the model using the following regularized squared error loss:
\begin{align}
\label{eq:ridge_regr}
\mathcal{L}_\lambda(\Gamma) := \frac{1}{N} \sum_{i=1}^{N} \left( y_i - \langle \Gamma, Z_i \rangle \right)^2 + \lambda \|\Gamma\|_F^2.
\end{align}
The solution to this optimization problem is denoted by $\Gamma^\star \in \mathbb{R}^{D \times D}$, whose vectorized form is given by
\begin{align}\label{eq:ridge}
\operatorname{vec}(\Gamma^\star; E_1,\ldots, E_N) = \bigg( \lambda N I + \sum_{k=1}^N \operatorname{vec}(Z_k) \operatorname{vec}(Z_k)^\top \bigg)^{-1} \sum_{k=1}^N y_{k, L+1} \operatorname{vec}(Z_k).
\end{align}
\begin{algorithm}[H]\caption{In-Context Differentially private pretraining of linear attention head (\fancyname)}\label{algo:DP2}
  \textbf{Input: }Training prompts $(E_k)_{k \in [N]}\in \R^{(D+1)\times (L+1)}$; noise scale $\sigma$; privacy parameters $\varepsilon, \delta$; clipping parameter $\mathcal{C}\geq 0$; projection parameters $R, G \geq 0$; regularization parameter $\lambda:=\lambda(n,d) \geq c>0$; number of iterations $T$; step-size $\eta_0$; and initialization $\Gamma^0 \in \R^{D \times D}$ with $\|\Gamma^0\|_F\leq R$.

  \begin{itemize}
      \item For $k\in [N]$, $\wt{Z}_k:= \Pi_G\left(L^{-1} x_{k, L+1} \sum_{i=1}^{L}\texttt{clip}_{\mathcal{C}}(y_{k,i}) x_{k,i}^\top\right)$. 
        \item For $t$ in $0,1, \ldots, T-1$:
        \begin{itemize}
            \item Generate $z_t \in \R^{D \times D}$ such that $\operatorname{vec}(z_t) \sim \dN_{D^2}\bigg(0, 2\eta_0^2 \frac{ T^2\sigma^2}{\varepsilon^2 N^2} \log\frac{1.25T}{\delta}\mathbb{I}_{D^2}\bigg)$.
            \item Do $\Gamma^{t+1} = \Pi_{R}\bigg((1-2\lambda \eta_0)\Gamma^t - \eta_0N^{-1}\sum_{k=1}^N \left(\langle \Gamma^t, \wt{Z}_k \rangle - \texttt{clip}(y_{k,L+1}) \right)\wt{Z}_k + z_t \bigg).$
        \end{itemize}
        \textbf{Output:} $\hat{\Gamma}:=\Gamma^T$.
  \end{itemize}
\end{algorithm}
\section{Differentially Private Pretraining}
\label{pre-train}
In this section, we present our differentially-private pretraining program of a linear attention network. Before proceeding to the main algorithm, we recall the definition of differential privacy. 

\begin{definition}
    A randomized algorithm $\mathcal{M}(\cdot)$ over a set of prompts is said to be in-context $(\varepsilon,\delta)$-differentially private if for any two sequences of prompts $\mathcal{D} = (E_1, \ldots, E_N)$ and $\mathcal{D}' = (E_1', \ldots, E_N')$ differing in at most one entry, 
    and for all measurable subsets $\mathcal{W}$ of outputs,
$$
\mathbb{P}[\mathcal{M}(\mathcal{D}) \in \mathcal{W}] \leq e^{\varepsilon} \mathbb{P}[\mathcal{M}(\mathcal{D}') \in \mathcal{W}] + \delta.
$$
\end{definition}
The probability is taken over the internal randomness of the mechanism $\mathcal{M}$, while the prompt sequences $\mathcal{D}$ and $\mathcal{D}'$ are treated as fixed. This definition ensures that the inclusion or exclusion of any individual data point has a limited effect on the algorithm’s output, thereby preserving privacy. A standard approach to enforce DP in the iterative training of machine learning models (e.g.\ gradient descent) is to inject noise at each update step. The cumulative effect of this noise is carefully calibrated to satisfy user-specified $(\varepsilon, \delta)$-differential privacy guarantees but minimize degradation in model performance. This technique, introduced as \emph{differentially private stochastic gradient descent}, has been echoed in recent works~\citep{abadi2016deep, tcai-cost-of-privacy, zhang2021understanding,gopi2021dp, majmudar2022differentially, bombari2025privacy}. In what follows, we improvise the aforementioned differentially-private training strategy while using the gradient descent to minimize the regularized loss $\mathcal{L}_\lambda(\Gamma)$ over a sequence of prompts: 
\[
\Gamma^{t+1} = (1 - 2\lambda \eta_0)\Gamma^t - \frac{\eta_0}{N} \sum_{k=1}^N \left( \langle \Gamma^t, Z_k \rangle - y_{k,L+1} \right) Z_k,
\]
where $\eta_0$ is the learning rate, and $\lambda$ is the regularization parameter.

To ensure privacy, we inject carefully calibrated Gaussian noise into each update step. The variance of this noise is set proportional to the \emph{$\ell_2$-sensitivity} of the update, which measures the maximum change in the update (in Frobenius norm) resulting from the change of a single training example. Formally, the $\ell_2$-sensitivity at iteration $t$ is defined as:
\begin{align}
\Delta(\hat{\Gamma}) = \left\| \hat{\Gamma}(E_1, \ldots, E_N) - \hat{\Gamma}(E'_1, \ldots, E'_N) \right\|_F,
\end{align}
where the datasets $(E_1, \ldots, E_N)$ and $(E'_1, \ldots, E'_N)$ differ in exactly one training prompt. 
Intuitively, privacy is preserved because an adversary observing the output of the algorithm (i.e., the final parameters) cannot reliably distinguish whether a change in the result is due to the presence or absence of a particular training prompt or due to the added random noise. 
However, in the problem setup considered in this paper, the $\ell_2$-sensitivity of the gradient updates may not be uniformly bounded across all possible sequences of training prompts due to the unbounded nature of the weights $w$ and the noise $\epsilon$. To mitigate this, we clip the responses $y_k$ and project the gradient updates $\Gamma^t$ onto compact sets. With these modifications, our differentially-private pretraining algorithm is presented in Algorithm~\ref{algo:DP2}, where the clipping and projection operators are defined as follows:
\[
\texttt{clip}_{\mathcal{C}}(x) := \arg\min_{y \in [-\mathcal{C}, \mathcal{C}]} \|x - y\|_2, \quad
\Pi_{R}(X) := \arg\min_{\substack{Y \in \mathbb{R}^{(D+1) \times (D+1)} \\ \|Y\|_F \leq R}} \|X - Y\|_F.
\]

\begin{theorem}\label{thm:DP-algo-2}
    Given the set of hyperparameters $(\mathcal{C}, R,G) \in \mathbb{R}^3_{>0}$, Algorithm~\ref{algo:DP2} is $(\varepsilon, \delta)$-differentially private if the noise scale $\sigma \geq 2G(\mathcal{C} + RG)$.
\end{theorem}
Theorem \ref{thm:DP-algo-2} (which we prove in Section \ref{pf_thm:DP-algo-2}) hints at the minimum amount of noise to be injected in the gradient descent step to achieve differential privacy. In particular, the amount of noise depends crucially on the projection parameters $\mathcal{C}$ and $R$. On the other hand, the higher the noise variance $\sigma^2$, the more we expect the predictive performance of the differentially-private estimate $\hat{\Gamma}$ to degrade compared to the ridge estimate $\Gamma^\star$ (as defined in \eqref{eq:ridge_regr}). However, it can still be argued that performing an appropriate number of iterations, governed by an ``early stopping criterion'', can improve accuracy. In fact, the additional error from noise injection can be made much smaller than the overall gradient descent error by properly tuning the hyper-parameters. This angle is explored in detail in Section \ref{se:price}.

\section{Cost of In-Context Differential Privacy}\label{se:price}
 In this section, we rigorously characterize the additional error incurred due to enforcing privacy constraints in Algorithm~\ref{algo:DP2}. Let $E^\texttt{test}$ be a test prompt and $y^\texttt{test}_{L+1}$ be the corresponding query response. Let us consider the prediction error in the test prompt given by
\[
\mathcal L_{\texttt{test}}(\Gamma) = (y^\texttt{test}_{L+1}-\langle \Gamma, Z(E^\texttt{test})\rangle)^2,
\]
where $Z(E^\texttt{test})$ is constructed from $E^\texttt{test}$ as described in \eqref{eq:construct_meta_varb}. We bound $\mathcal L_{\texttt{test}}(\Gamma)$ by the following two types of error terms:
\[
\mathcal L_{\texttt{test}}(\hat{\Gamma}) \leq 2\mathcal L_{\texttt{test}}(\Gamma^\star) + 2(\langle \hat{\Gamma}, Z(E^\texttt{test})\rangle-\langle \Gamma^\star, Z(E^\texttt{test})\rangle)^2.
\]
While $\mathcal L_{\texttt{test}}(\Gamma^\star)$ is the prediction error of the non-private procedure, the extra error is proportional to $(\langle \hat{\Gamma}, Z(E^\texttt{test})\rangle-\langle \Gamma^\star, Z(E^\texttt{test})\rangle)^2$. The following theorem characterizes this extra error.

\begin{theorem}
    \label{thm:main_thm}
Consider the pretrained weights $\hat{\Gamma}$ generated by running \fancyname\, (Algorithm \ref{algo:DP2}) on prompt set $(E_1, \ldots, E_N)$ , ensuring $(\varepsilon,\delta)$ differential privacy for $T$ iterations with a fixed stepsize $\eta_0 \in (\frac{\lambda}{c(2\lambda + G^2)^2}, \frac{\lambda}{(2\lambda + G^2)^2})$ for some large $c>1$, and $\Gamma^\star$ generated by solving the ridge regression described in \eqref{eq:ridge_regr}. If the clipping and projection parameters are set as:
\begin{align}
    &\nu=1+\tau^2, \quad \mathcal C =\sqrt{2\nu\log(NL/\kappa)}, \quad  G=\frac{\mathcal C}{\sqrt L}\Big(1+\frac{(\log (N/\kappa))^{1/2}}{D}\Big),\nonumber\\
    & G_0 = \frac{\mathcal C}{\sqrt L}\Big(1+\frac{(\log (1/\kappa))^{1/2}}{D}\Big), \quad \text{and } R \asymp \lambda^{-1}\mathcal{C}^2\sqrt{\frac{N}{L}} \Big(1+\frac{(\log(1/\kappa))^{1/2}}{D}\Big),\label{eq:hyperparameter-choice}
\end{align}
then for a test prompt $E$ independent of $(E_k)_{k\in[N]}$,
\begin{align}
    (\langle \hat{\Gamma} , Z\rangle  - \langle \Gamma^{\star}, Z\rangle)^2 \leq G_0^2 \bigg( (1-\eta_0\lambda)^{T} R^2 + \sigma^2 \eta_0^2 D \frac{T^2 \log(2T/\delta)}{N^2 \varepsilon^2}\bigg).\label{eq:hp-bound-final}
\end{align}
with probability greater than $1 - c_1 \exp(-c_2 D) -  4\kappa$, where $Z$ is formed via $E$ as in \eqref{eq:construct_meta_varb}.
\end{theorem}
The above theorem is proved in Section \ref{pf_thm:main_thm}.
\begin{remark} \label{remark-following-main-thm}
    The ``\textit{cost of privacy}'' on the right hand side of \eqref{eq:hp-bound-final} naturally decomposes into two components. The first arises from the optimization error of gradient descent, hereby referred to as the ``\textit{cost of descent}'', and is given by $(1 - \eta_0 \lambda)^T$, where $\eta_0$ is the step size, $\lambda$ is the strong convexity parameter, and $T$ is the number of gradient steps. The second component stems from the noise injected at each iteration to ensure DP, and takes the form
$
\sigma^2 \eta_0^2 D \cdot \frac{T^2 \log(2T/\delta)}{N^2 \varepsilon^2},
$
where $\sigma^2$ denotes the variance of the added noise, $D$ is the feature dimension, $N$ is the number of training samples, and $(\varepsilon, \delta)$ are the privacy parameters. This term will henceforth be referred to as the ``\textit{cost of noise injection}''. The trade-off between these two terms plays a crucial role in determining the generalization error. While the optimization error decays exponentially with $T$, the privacy-induced error increases quadratically. Therefore, it is essential to choose an optimal stopping time for the gradient descent iterations. This optimal stopping time depends on the problem hyper-parameters $\eta_0$, $\lambda$, and the feature dimension $D$. In the following theorem (proved in Section \ref{pf_cor:dic_behav}), we characterize how the interplay between the dimensionality and the stopping time governs the behavior of the generalization error in different settings of interest.
\end{remark}
\begin{theorem}
\label{cor:dic_behav}
 Assume $N \gtrsim D^2 L^{-2}$, and suppose the noise scale $\sigma$ in Algorithm \ref{algo:DP2} satisfies $\sigma \asymp 2G(\mathcal{C} + RG)$ where $(\mathcal{C}, R, G)$ is the set of hyper-parameters. Then, under the assumptions and hyper-parameter specifications of Theorem \ref{thm:main_thm}, the following assertions hold:
 \begin{enumerate}[label=(\roman*)]
     \item (Low-dimensional setting) If $\kappa \gtrsim \exp(-D^2)$, and $D^2 \lesssim \log(NL)$, then, after $T=\frac{\log (N^2 L D^3)}{\log ((1-\eta_0\lambda)^{-1})}$ many iterations of Algorithm \ref{algo:DP2} with $\eta_0 \asymp \lambda \asymp 1$ such that $\eta_0\lambda \in (0,1)$, the cost of privacy of $\hat{\Gamma}(=\Gamma^{T})$ behaves as follows:
     \begin{align}\label{eq:low-dim}
         (\langle \hat{\Gamma} , Z\rangle  - \langle \Gamma^{\star}, Z\rangle)^2 \lesssim \nu^5 \frac{\log^{10} (NL)}{NL^3}\left(1+  \frac{\log(1/\delta)}{\varepsilon^2}\right),
     \end{align}
      with probability at least $1 - c_1 \exp(-c_2 D) -  4\kappa$.
\item (High-dimensional setting) If $\kappa \gtrsim (NL)^{-1}$, and $D^2 \gtrsim \log(NL)$, then for some $r$ (possibly depending on $N,L$ and $D$), let $T=\frac{\log r}{\log((1-\eta_0\lambda)^{-1})}$. If $\eta_0< \frac{\lambda}{(2\lambda+G^2)^2}$, then
     \begin{align}\label{eq:high-dim}
         &(\langle \hat{\Gamma} , Z\rangle  - \langle \Gamma^{\star}, Z\rangle)^2 \nonumber\\
         &\lesssim \frac{N\nu^5 \log^3(NL)}{L^2 \lambda^2 r} \bigg(1 + \frac{D \ r\log^3r}{N^3} \left(1 + \log^2(NL){\frac{N}{L^2\lambda^2}}\right)\frac{\log(1/\delta)}{\varepsilon^2}\bigg),
     \end{align}
      with probability at least $1 - c_1 \exp(-c_2 D) -  4\kappa$.
 \end{enumerate}
\end{theorem}
\begin{remark}
\label{rem:opt_stop}
  In the low-dimensional setting with a specific choice of $T$, the cost of gradient descent is negligible, and the cost of noise injection becomes the dominant contributor to the overall cost of privacy. Notably, the restriction on $D$ renders it irrelevant in determining the cost of privacy in this regime. On the other hand, among the many possible high-dimensional scenarios, a particularly interesting case is the over-parameterized regime where $N \asymp L^2 \asymp D^2$.\end{remark}
  
\begin{proposition} \label{prop1}
    If $N \asymp L^2 \asymp D^2$, then with $\lambda \asymp \frac{N}{D}$, it holds that 
    \begin{align} \label{eq:over-parametrize}
    \big( \langle \hat{\Gamma} , Z\rangle  - \langle \Gamma^{\star}, Z\rangle \big)^2 
    \lesssim \nu^5 \log^3(NL) \frac{D^2}{NL^2 r} \Big(1 + r\log^3 r \cdot \frac{D}{N^3} \cdot \frac{\log(1/\delta)}{\varepsilon^2} \Big),
\end{align}
with probability at least $1 - c_1 \exp(-c_2 D) -4\kappa$. In particular, when $r \asymp N$, or, equivalently, $T\asymp\log N,$ it holds that 
\begin{align}\label{eq:early-stopping}
    \big( \langle \hat{\Gamma} , Z\rangle  - \langle \Gamma^{\star}, Z\rangle \big)^2 
    \lesssim \nu^5 \log^3(NL) \frac{D^2}{N^2L^2}
\end{align}
    with probability at least $1 - c_1 \exp(-c_2 D) -  4\kappa$.
\end{proposition}
  This result is proved in Section \ref{pf_prop1}. The choice of $\lambda$ in Proposition~\ref{prop1} is standard and also appears in the ridge regression analysis of \cite{lu2024context}. Equation~\eqref{eq:over-parametrize} highlights the trade-off between the two components of the cost of privacy, as previously discussed in Remark~\ref{remark-following-main-thm}. For fixed values of $N$, $L$, and $D$, the test risk of the estimates generated by \fancyname\, decreases with the number of iterations $T$ at a rate of $\Theta(e^{-T})$ whereas the test error increases at a rate of $\Theta(T^3)$. 
  As a result, in this high-dimensional regime, the \emph{optimal stopping point} for pretraining is $T = \Theta(\log N)$ iterations. 
  This phenomena is explored numerically in Section \ref{se:N vs eps}.

\section{Robustness properties of \fancyname } \label{se:robustness}
 In this section, we demonstrate that \fancyname{} is inherently robust to adversarial perturbations to the training data. Specifically, we show that such perturbations during the pretraining stage affect the generalization error of our method significantly less than the baseline approach proposed in~\cite{lu2024context}.

Consider a set of training prompts $E_1, \ldots, E_N$, and suppose a malicious attacker aims to degrade performance on an independent test prompt $E$ by perturbing the training data, thereby inducing inaccurate estimation of the weights in the attention module. To disrupt the training process, the attacker selects a prompt uniformly at random from the training set, say $E_i$, and replaces it with a perturbed version,
\begin{align} \label{eq:prompt-perturbation}
    E_{\mathrm{bad}, i}(\mu, \alpha) = \begin{pmatrix}
        x_{i,1}' & x_{i,2}' & \cdots & x_{i,L}' & x_{i,L+1}' \\
        y_{i,1}' & y_{i,2}' & \cdots & y_{i,L}' & 0
    \end{pmatrix} \in \mathbb{R}^{(D+1) \times (L+1)},
\end{align}
where the perturbed components are given by $x_{i,k}' = x_{i,k} + \mu$ for all $k \in [L+1]$ and $y_{i,\ell}' = y_{i,\ell} + \alpha$ for all $\ell \in [L]$. Let the parameter trained by the \fancyname~algorithm acting on the perturbed set of prompts $(E_1, \ldots, E_{i-1}, E_{\bad,i}, E_{i+1}, \ldots, E_N)$ be $\hat{\Gamma}_\bad$. Correspondingly, let the parameter trained on the original, unperturbed prompts be $\hat{\Gamma}$. Let the ridge regression solutions of \eqref{eq:ridge} on the ``perturbed'' and ``original'' set of prompts, be denoted by $\Gamma^{\star}_\bad$ and $\Gamma^{\star}$, respectively. Then the following theorem characterizes the robustness properties of the estimates generated by \fancyname.
\begin{theorem}\label{thm:robustness-thm}
    Consider the \fancyname\ algorithm with the hyper-parameter specifications as in Theorem \ref{thm:main_thm}. Further, consider an adversarial prompt perturbation as in \eqref{eq:prompt-perturbation}, with $\mu,\alpha$ satisfying 
\begin{align}
    & \alpha^2 \mu^4 \leq c_u NL\lambda \quad \mbox{and} \quad \alpha^2\mu^2 \geq c_\ell \mathcal{C}^2 L^{-1/2} (1 \vee \lambda NR^2L^{-1/2}) \label{eq:lowerbound},
\end{align}
for large enough constant $c_u>0$ and small enough constant $c_\ell>0$. If $\kappa>Ne^{-D^2}$ and $\lambda > \mathcal{C}^2L^{-1}$, then for an ``unperturbed'' test prompt $E$ and the corresponding $Z$ from \eqref{eq:construct_meta_varb}, it holds that
\begin{align} \label{eq:robustness-bound}
      (\langle \hat{\Gamma} , Z\rangle  - \langle \hat{\Gamma}_\bad, Z\rangle)^2 \lesssim \frac{N}{L^2}\log^2(NL/\kappa) <  \frac{\alpha^2\mu^2}{N\lambda}\leq (\langle \Gamma^\star , Z\rangle  - \langle \text{$\Gamma^{\star}_\bad$}, Z\rangle)^2,
\end{align}
with probability at least $1 - c_1\exp(-c_2D)-5\kappa$ for constants $c_1, c_2>0$.
\end{theorem}
\begin{remark}
    Theorem~\ref{thm:robustness-thm} (proved in Section~\ref{pf_thm:robustness-thm}) shows that under bounded perturbations, pretraining with \fancyname{} yields generalization error closer to that from the unperturbed setup than does ridge regression. If $\lambda \asymp 1$ and $D^2 \gtrsim \log N$, the bounds in \eqref{eq:lowerbound} simplify to $\frac{N^2}{L^2} \lesssim \alpha^2\mu^2 \leq \alpha^2\mu^4 \lesssim NL$. In the regime $\frac{N}{L^2} \log^2(NL) \to 0$, an adversary can choose $\alpha, \mu$ such that $\alpha^2\mu^2 \to \infty$ while still satisfying $\alpha^2 \mu^4 \lesssim NL$, leading to $(\langle \Gamma^\star , Z\rangle - \langle \Gamma^\star_\bad, Z\rangle)^2 \xrightarrow{\IP} \infty$. In contrast, \fancyname~ ensures $(\langle \hat{\Gamma} , Z\rangle - \langle \hat{\Gamma}_\bad, Z\rangle)^2 \xrightarrow{\IP} 0$ even under such adversarial conditions, as confirmed by experiments in Section~\ref{se:robustness-simu}.
\end{remark}
\section{Numerical experiments} \label{se:simulation}
We evaluate the empirical behavior of the \fancyname{} algorithm. Section~\ref{se:N vs eps} examines how prediction risk changes under different privacy constraint strengths. Section~\ref{se:over-parametrized} explores the trade-off between optimization and noise under different iteration counts. Section~\ref{se:robustness-simu} validates the robustness of \fancyname\ to adversarial perturbations. All code to reproduce the figures can be found at \href{https://github.com/kingsleyyeon/DP}{https://github.com/kingsleyyeon/DP}.
\begin{figure}[H]
  \centering
  \begin{minipage}[t]{0.48\textwidth}
    \centering
    \includegraphics[width=0.95\textwidth]{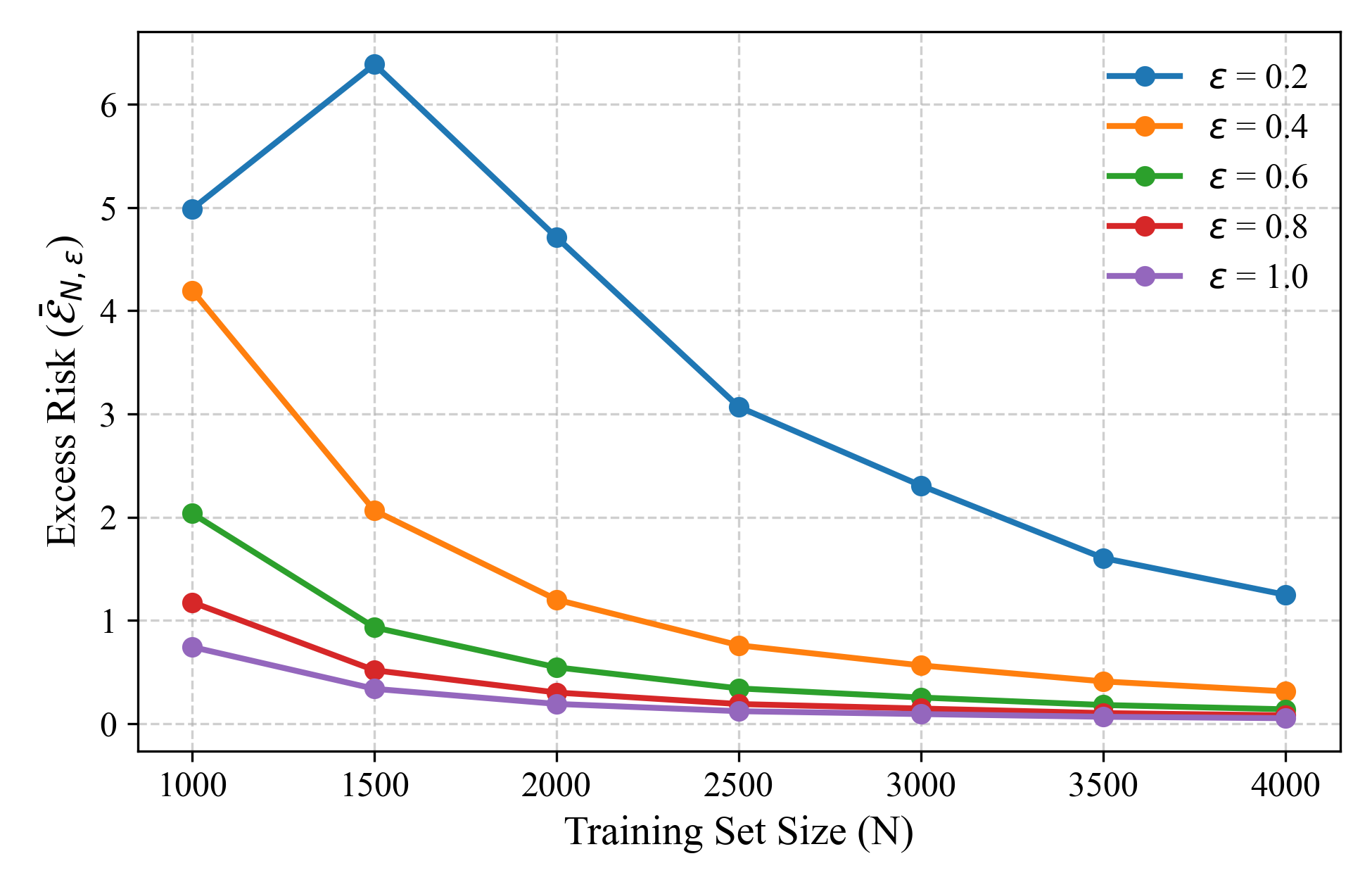}
    \caption{Excess risk of \fancyname\ for the low-dimensional set-up with $D=5$.}
    \label{fig:low_dim_vs_N_eps_left}
  \end{minipage}
  \hfill
  \begin{minipage}[t]{0.48\textwidth}
    \centering
    \includegraphics[width=\textwidth]{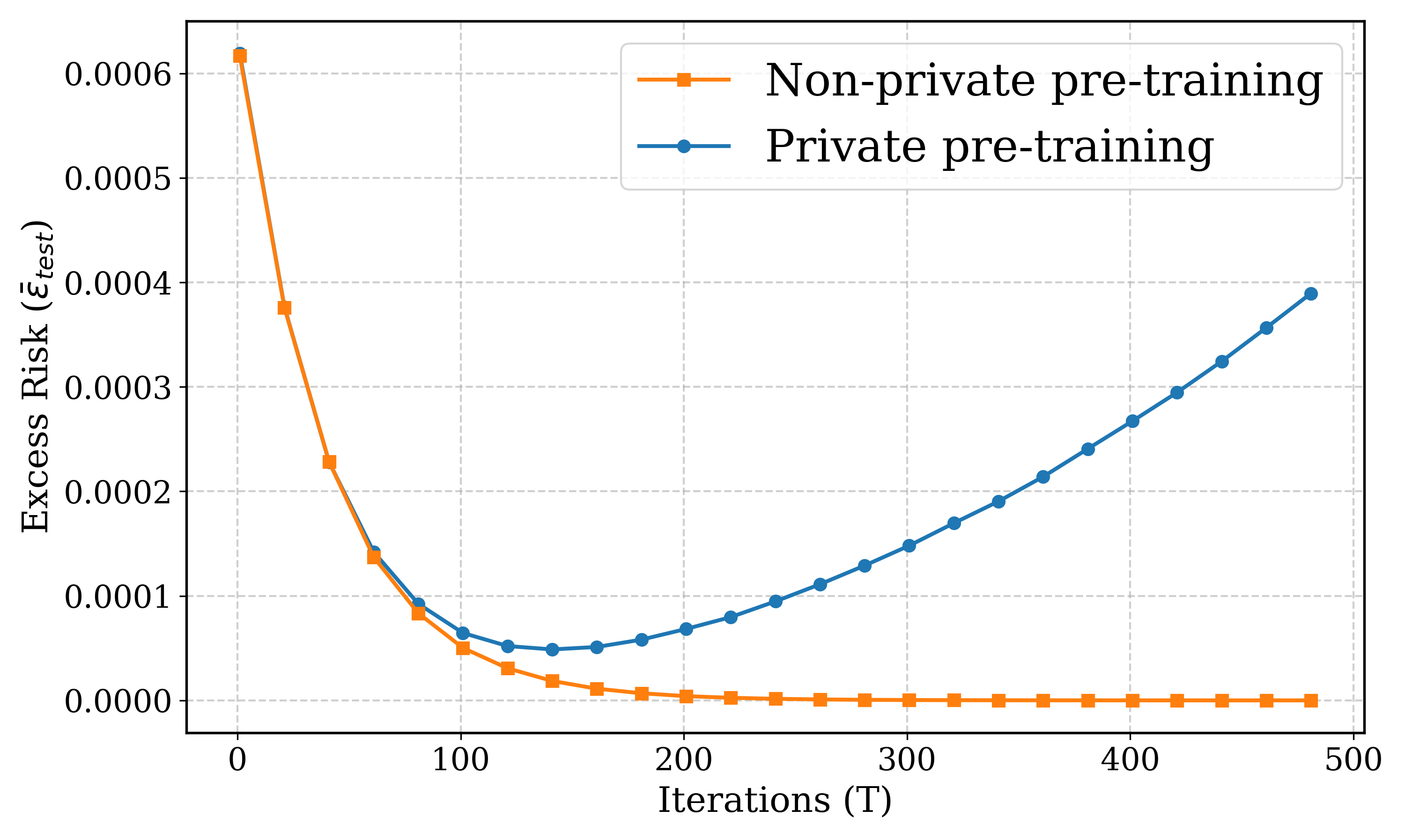}
    \caption{Interplay between the cost of descent and the cost of privacy in the overparameterized setting with $N=1000$ and $\varepsilon=0.8$.}
    \label{fig:phase-transition}
  \end{minipage}
\end{figure}
\subsection{Effect of privacy on prediction risk} \label{se:N vs eps}
We evaluate the impact of privacy on ICL in a low-dimensional setting with $D = 5$. Full simulation details, as well as high-dimensional experiments, are provided in Appendix~\ref{se:N vs eps}. In this experiment, we vary the number of prompts $N$ and privacy level $\varepsilon$. We use $T = \log N^{5/2} / \log(1 - \lambda \eta_0)$ and set other parameters according to Theorem~\ref{thm:main_thm}. The excess test risk, averaged over $B = 500$ trials, is measured relative to that of ridge regression as $\mathcal{E}_{\operatorname{test}} = \frac{1}{n_{\operatorname{test}}} \sum_{k=1}^{n_{\operatorname{test}}} \big(\langle \hat{\Gamma} - \Gamma^\star, Z_{k,\operatorname{test}} \rangle \big)^2$, where $\hat{\Gamma}$ is the DP estimate and $\Gamma^\star$ is the ridge regression solution. As shown in Figure~\ref{fig:low_dim_vs_N_eps_left}, the excess risk decreases with $N$ and increases under stricter privacy, aligning with Theorem~\ref{cor:dic_behav}.

\subsection{Effect of early stopping in over-parametrized setting} \label{se:over-parametrized}
Next, we fix $N=1000$ and study how the test error evolves with the number of iterations $T$, in an over-parameterized regime with $L \asymp D \asymp \sqrt{N}$ and $\varepsilon = 0.8$. For each $T$, we average test error over $500$ trials, comparing that of the differentially-private training with that of non-private gradient descent. Figure~\ref{fig:phase-transition} illustrates a phase transition in DP training: first the error decreases with $T$ as optimization improves the solution, but then it passes a critical point and the error rises as injected noise accumulates. Early stopping around $T \approx 140$ optimally balances under-optimization and noise accumulation in this setting. This validates the need for early stopping under privacy constraints.
\begin{figure}[H]
  \centering
  \begin{minipage}[t]{0.48\textwidth}
    \centering
    \includegraphics[width=\textwidth]{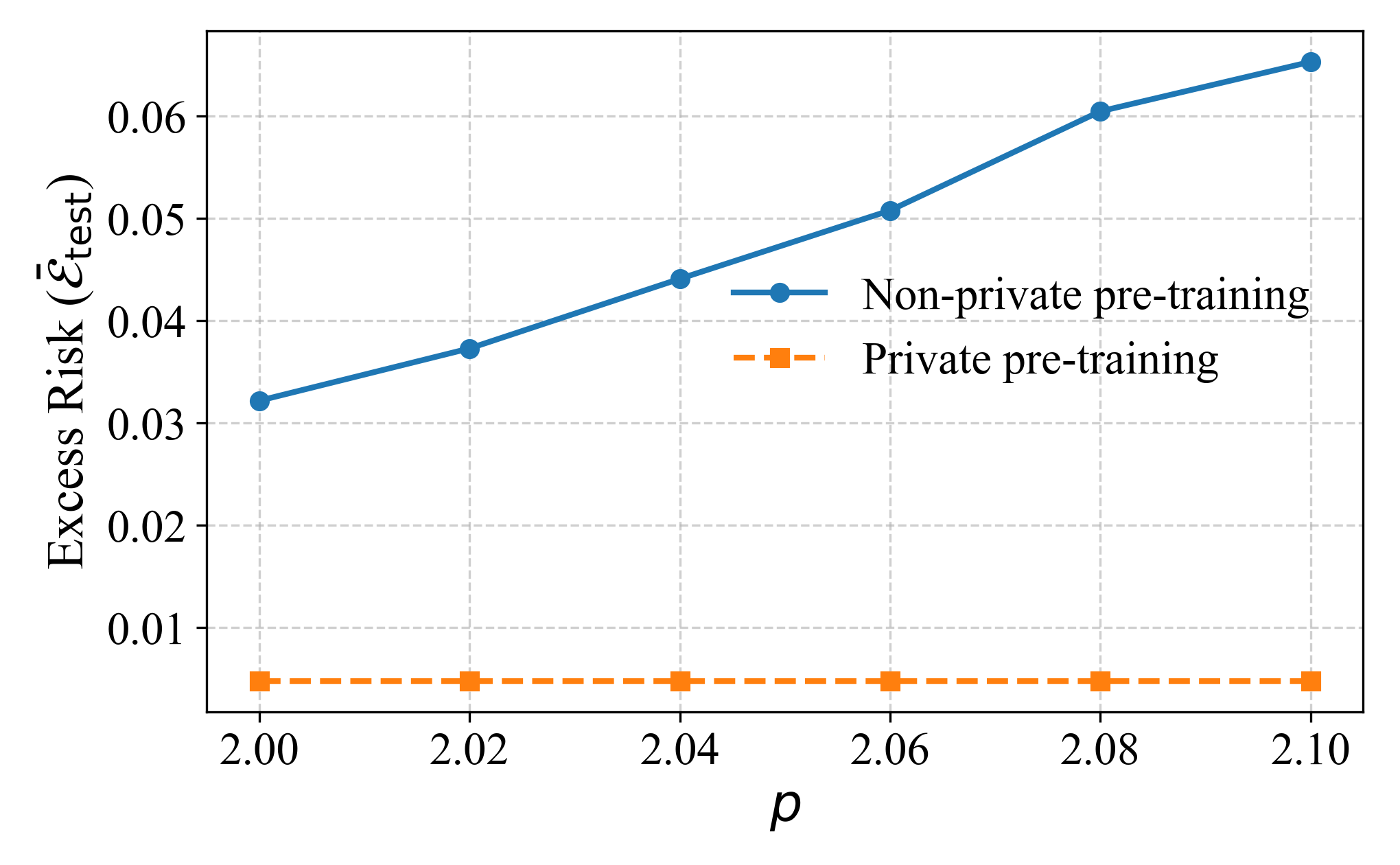}
  \end{minipage}
  \hfill
  \begin{minipage}[t]{0.48\textwidth}
    \centering
    \includegraphics[width=\textwidth]{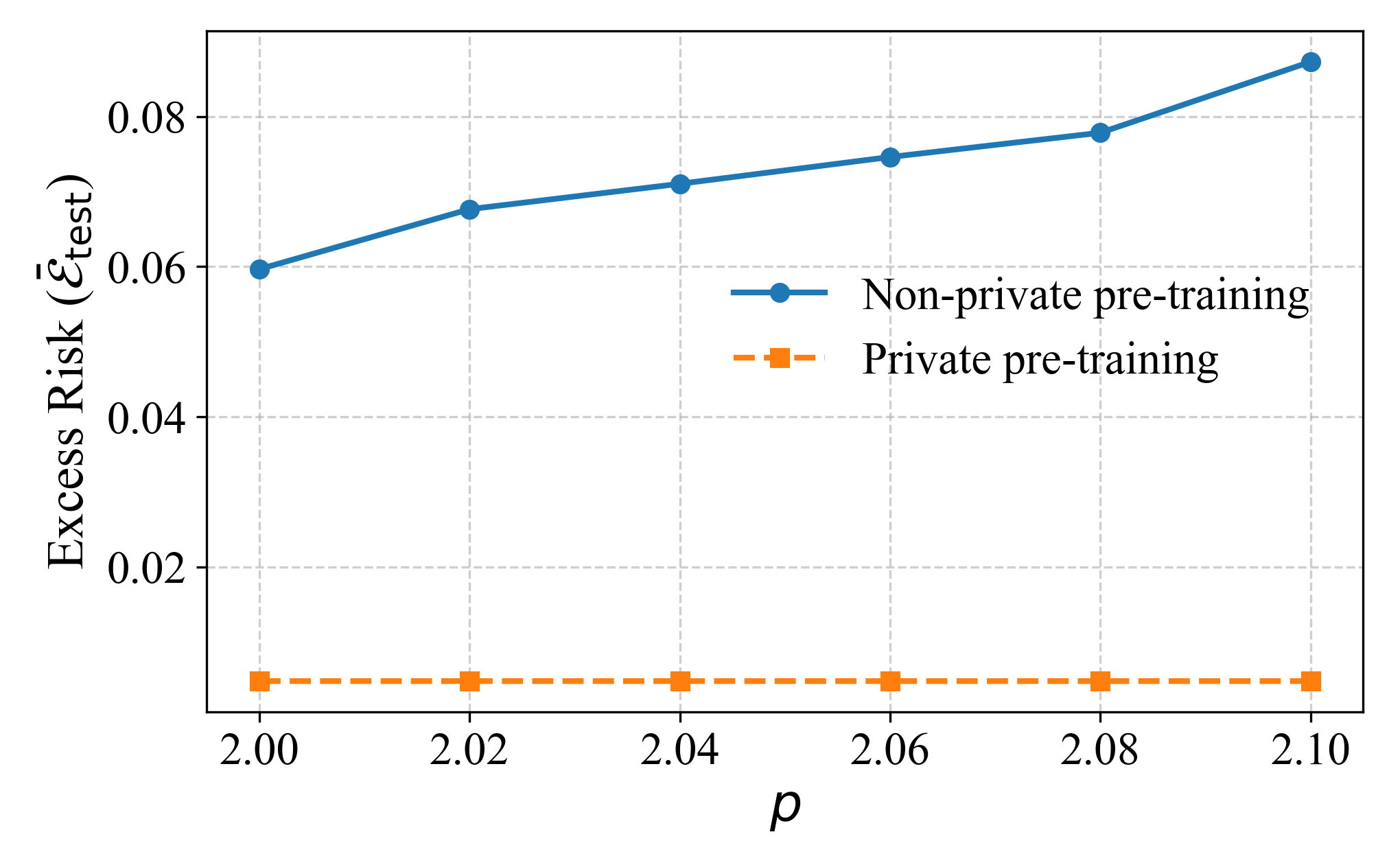}
  \end{minipage}
  \caption{Comparison of prediction error under adversarial perturbations for different values of $c$. Left: $c = 2$; Right: $c = 4$. The differentially private estimator (\fancyname) consistently outperforms the ridge estimator ($\Gamma^\star$) as the perturbation magnitude $\alpha = cN^p$ increases.}
  \label{fig:risks_comparison}
\end{figure}

\subsection{Robustness of \fancyname} \label{se:robustness-simu}

We test robustness by changing one training prompt with the additive perturbation $\alpha = cN^p$ (as described in Section~\ref{se:robustness}) for $c \in \{2, 4\}$ and $p \in [2, 2.1]$, while fixing $N = 5000$, $L = 500$, and $D = 5$. We compare the prediction error of \fancyname\ and ridge regression on $500$ test prompts across $500$ trials. Figure~\ref{fig:risks_comparison} shows that while ridge regression is increasingly affected by larger perturbations, \fancyname\ remains robust, demonstrating its resilience to adversarial training examples.

\section{Conclusion}
Maintaining privacy during the pretraining of large language models is an increasingly important challenge as such architectures become ubiquitous. To the best of our knowledge, this work provides the first systematic theoretical characterization of differentially-private in-context learning. We quantify the \emph{cost of privacy} on the performance of linear attention heads and formally justify the widely observed phenomenon of \emph{early stopping}~\citep{zhangearlystopping, majmudar2022differentially, bu2024pre, bombari2025privacy} in the context of training attention-based models under differential privacy—previously unexplored even for simplified architectures using the attention mechanism. Recent studies~\citep{dai2023transformers, vladymyrov2024linear, liang2025transformers} show that multi-layered transformers can emulate gradient-based learning. Our framework offers a pathway toward understanding the theoretical behavior of such models when executing privacy-preserving pretraining, with potential implications for mitigating the ``regurgitation'' \cite{carlini2021extracting} behavior observed in large language models. 

\bibliographystyle{abbrvnat}
\bibliography{references}

\appendix

\section{Theoretical Details}

\subsection{Choice of parameters} \label{se:choice}
\cite{yu2023dptraining} provides an intuitive explanation for the behavior of the predictor \eqref{eq:yhat_old}. The term $w^{PV}_{22} W^{KQ}_{11}$ is approximately equal to $\mathbb{E}[(X^\top X)^{-1}]$, capturing the inverse second-moment structure of the features. The second term does not depend on the features, and the third term is independent of the labels $y$. They act as an extra additive term, which can be assumed to have no significant impact on the final prediction. The fourth term represents the effect of projecting the input features $x_i$ onto the direction $w^{PV}_{12}$ in the final prediction. However, since the features are assumed to be isotropic, it is reasonable to expect that projections onto any particular direction carry no special predictive value. Consequently, it is justified to assume that $w^{KQ}_{12}=0$ and $w^{PV}_{12} = 0$, which simplifies the predictor to:
\[
\hat{y}_{L+1} = \frac{1}{L} \left\langle x_{L+1}, w^{PV}_{22} W^{KQ}_{11} \sum_{i=1}^L y_i x_i \right\rangle.
\]

This assumption is further supported by the observation in \cite{bartlett-icl-lm-24}, where the authors show that when the parameters of $W^{PV}$ and $W^{KQ}$ are learned via gradient flow on the average reconstruction loss $\mathbb{E}[(\widehat{y} - y)^2]$, initializing with $w^{KQ}_{12}=0$ and $w^{PV}_{12} = 0$ ensures that the parameter remains zero throughout training.

\subsection{Proof of Theorem \ref{thm:DP-algo-2}}
\label{pf_thm:DP-algo-2}
    Consider two datasets of prompts $(E_k)_{k \in [N]}$ and $(E'_k)_{k \in [N]}$ that differ in exactly one prompt. Without loss of generality, assume $E_1 \neq E'_1$ and $E_k = E'_k$ for all $k \geq 2$. The change in the gradient update due to this single difference is bounded by:
    \begin{align}
        &\frac{\eta_0}{N} \bigg( \|\langle \hat{\Gamma}, \widetilde{Z}_1 \rangle \widetilde{Z}_1\|_F + \|\langle \hat{\Gamma}, \widetilde{Z}_1' \rangle \widetilde{Z}_1'\|_F + \|\texttt{clip}_{\mathcal{C}}(y_{1,L+1}) \widetilde{Z}_1\|_F + \|\texttt{clip}_{\mathcal{C}}(y_{1,L+1}') \widetilde{Z}_1'\|_F \bigg) \nonumber \\
        &\leq \frac{2\eta_0(R G^2 + \mathcal{C}G)}{N} \leq \frac{\eta_0 \sigma}{N}, \label{eq:bound-sensitivity}
    \end{align}
    where the final inequality follows from the assumption on $\sigma$.

    By Lemma 2.5 of \citet{kamath2020primer}, each gradient step in Algorithm~\ref{algo:DP2} is $(\varepsilon/T, \delta/T)$-differentially private. The overall guarantee then follows by composition, using Fact 2.2 of \citet{tcai-cost-of-privacy}.

\subsection{Proof of Theorem \ref{thm:main_thm}}
\label{pf_thm:main_thm}
 Consider the set $\mathcal{D}_1:=\{\| \Gamma^{\star}\|_F \leq R\} $. Moreover, denote 
    \[ \wt{\Gamma}^{t+1}=(1-2\lambda)\Gamma^t - \eta_0N^{-1}\sum_{k=1}^N \left(\langle \Gamma^t, \wt{Z}_k \rangle - \texttt{clip}(y_{L+1}) \right)\wt{Z}_k .\]
    Clearly, $\Gamma^{t+1}=\Pi_R(\wt{\Gamma}^{t+1} + \boldsymbol{z}_t)$. Under $\mathcal{D}_1$, it is easy to see that \begin{align}
       \|\hat{\Gamma} - \Gamma^{\star} \|_F^2 &\leq \|\wt{\Gamma}^T + \boldsymbol{z}_{T-1} - \Gamma^{\star}\|_F^2 \leq  (1+C_0^{-1}) \|\wt{\Gamma}^T - \Gamma^{\star} \|_F^2 + (1+C_0) \| \boldsymbol{z}_{T-1}\|_F^2,  \label{eq:decomp-1}
    \end{align}
    where, the choice of the constant $C_0$ ensures\begin{align} \label{eq:kappa}
        (1+C_0^{-1}) \kappa <1-\eta_0\lambda \text{ , with } \kappa:= 1 - 2\eta_0 \lambda + \eta_0^2(G^2+2\lambda)^2.
    \end{align} Further consider the sets $\mathcal{D}_2:= \left\{\max_{k\in [N]} \left\|\sum_{i=1}^L x_{k,i}\right\|_2 \leq G L\mathcal{C}^{-1} \right\}$, and $\mathcal{D}_3:=\left\{ \max_{k\in [N], i \in [L+1]} |y_{k,i}| \leq \mathcal{C} \right\}$. Since $\|x_{k,L+1}\|_2=1$, under the events $\mathcal{D}_2$ and $\mathcal{D}_3$, it follows that 
    \begin{align}\label{eq:bound-on-unprojected-Z}
        \max_{k \in [N]}\left\|L^{-1} x_{k, L+1} \sum_{i=1}^{L}y_{k,i} x_{k,i}^\top\right\|_F \leq G ,
    \end{align}
    which implies $\wt{Z}_k = Z_k$ for all $k \in [N]$ by the definition of $Z_k$ in \eqref{eq:construct_meta_varb}.
    The sets $\mathcal{D}_i, i=1,2,3$ allow us to bear down the classical theory of convex minimization, and our choice of the parameters $R, G$ and $\mathcal{C}$ will emphasize that these events occur with high probability. In particular, under $\mathcal{D}:=\cap_{i=1}^3\mathcal{D}_i$, we note the $\mathcal{L}$ is $\lambda$-strongly convex:
  \begin{align} \label{eq:strong-convexity}
      \langle \nabla_{\Gamma} \mathcal{L}(\Gamma, (Z_k)_{k \in [N]}) , \Gamma - \Gamma^{\star}\rangle \geq \lambda \|\Gamma - \Gamma^{\star} \|_F^2 ,
  \end{align}
  and the $(G^2+2\lambda)$-smooth: \begin{align}\label{eq:lipschitz}
      \left\|\nabla_{\Gamma}\mathcal{L}(\Gamma, (Z_k)_{k \in [N]}) - \nabla_{\Gamma}\mathcal{L}(\Gamma', (Z_k)_{k \in [N]}) \right\|_F \leq (G^2+2\lambda)\, \|\Gamma - \Gamma' \|_F.
  \end{align}
Therefore, for the term $\|\wt{\Gamma}^T - \Gamma^{\star}\|_F$ in \eqref{eq:decomp-1},
\begin{align}
    \|\wt{\Gamma}^T - \Gamma^{\star} \|_F^2 = & \| \Gamma^{T-1} - \eta_0 \nabla_{\Gamma^{T-1}} \mathcal{L}(\Gamma^{T-1}, (Z_k)_{k \in [N]} ) - \Gamma^{\star}\|_F^2\leq  \kappa \|\Gamma^{T-1} - \Gamma^{\star} \|_F^2 , \label{eq:gd-analysis}
\end{align}
    where we recall $\kappa$ from \eqref{eq:kappa}, and \eqref{eq:gd-analysis} employs \eqref{eq:strong-convexity} and \eqref{eq:lipschitz}. Note that we must require $\kappa<1$, which makes use of $\eta_0 < \frac{\lambda}{(G^2 + 2\lambda)^2}$. Putting \eqref{eq:gd-analysis} back into \eqref{eq:decomp-1}, one obtains under $\mathcal{D}$ that
    \begin{align*}
         \|\hat{\Gamma} - \Gamma^{\star} \|_F^2&\leq (1+C_0^{-1})\kappa\| \Gamma^{T-1} - \Gamma^{\star}\|_F^2 + (1+C_0) \| \boldsymbol{z}_{T-1}\|_F^2.
    \end{align*}
    Proceeding recursively, we can show that for all $T>1$, we have
    \begin{align}
    \|\hat{\Gamma} - \Gamma^{\star} \|_F^2 \leq (1-\eta_0\lambda)^TR^2 + (1+C_0) \sum_{i=0}^{T-1} (1-\eta_0\lambda)^{T-i-1} \|\boldsymbol{z}_{i}\|_F^2.
    \label{eq:noise-sum}
    \end{align}
    Since the errors $(\boldsymbol{z}_i)_{i=1}^T$ are independent of the prompts $(E_k)_{k \in [B]}$, an application of Lemma A.2. of \cite{tcai-cost-of-privacy} implies
    \begin{align*} 
        \|\hat{\Gamma} - \Gamma^{\star} \|_F^2 \lesssim (1-\eta_0 \lambda)^{T} R^2 + \sigma^2 \eta_0^2 D \frac{T^2 \log(2T/\delta)}{N^2 \varepsilon^2}, 
    \end{align*}
   with probability at least $1 - c_1 \exp(-c_2 D)$ under $\mathcal{D}$. An application of Cauchy-Schwarz inequality entails
   \begin{align}\label{eq:hp-conditional}
       (\langle \hat{\Gamma} , Z\rangle  - \langle \Gamma^{\star}, Z\rangle)^2 \leq G_0^2 \bigg( (1-\eta_0\lambda)^{T} R^2 + \sigma^2 \eta_0^2 D \frac{T^2 \log(2T/\delta)}{N^2 \varepsilon^2}\bigg)
   \end{align}
    with probability at least $1 - c_1 \exp(-c_2 D)$ under $\mathcal{D} \cap \{\|Z\|_F \leq G_0\}$. Now we turn to tackling the individual events $\mathcal{D}_i$, $i=1,2,3$. For $\mathcal{D}_3$, note that if $(x_i)_{i\in [L]} \overset{i.i.d.}{\sim} \mathcal{U}(\mathbb{S}^{D-1})$ and $w \sim N(0, \mathbb{I}_D)$ independently of $x_i$'s, then $(w^\top x_i){\sim} N(0, 1)$ marginally. Therefore, Lemma \ref{lemma:C-term} implies 
    \begin{align}\label{eq:D3}
        \IP(\mathcal{D}_3) \geq 1- \kappa, \quad \text{ for $\mathcal C =\sqrt{2\nu\log(NL/\kappa)}$.}
    \end{align}
    Furthermore, with $G \asymp \frac{\mathcal C}{\sqrt L}\left(1+\left(\frac{\log (N/\kappa)}{D^2}\right)^{1/4}\right)$, from Lemma \ref{lemma:G-term} we get that 
    \begin{align}\label{eq:D2}
        \IP(\mathcal{D}_2) \geq 1- \kappa.
    \end{align}
    Finally, noting that 
    \begin{align*}
        \sum_{k=1}^N y_{k, L+1}\operatorname{vec}(Z_k) \leq  \max_{k\in [N]} |y_{k, L+1}| \left(\max_{k\in [N], i \in [L]} |y_{k,i}|\right) \frac{1}{L} \sum_{k,i} \operatorname{vec}(x_{k,L+1}x_{k,i}^\top),
    \end{align*}
    an application of Lemma \ref{lemma:C-term} on \eqref{eq:ridge}, in conjunction with Lemma \ref{lemma:R-term}, yields 
    \begin{align}\label{eq:D1}
        \IP(\mathcal{D}_1)\geq 1 - \kappa, \text{ with } R= \lambda^{-1}\mathcal{C}^2\sqrt{\frac{N}{L}} \left(1+\left(\frac{\log(1/\kappa)}{D^2}\right)^{1/4}\right).
    \end{align}
    Finally, similar to Lemma \ref{lemma:G-term} it can be argued that 
    \begin{align}
    \label{eq:bound_indv_z}
    \IP(\|Z\|_F \leq G_0) \geq 1- \kappa, \quad  \mbox{for $G_0 \asymp \frac{\mathcal C}{\sqrt L}\left(1+\left(\frac{\log (1/\kappa)}{D^2}\right)^{1/4}\right)$.}
    \end{align} 
    Summarizing \eqref{eq:D3}-\eqref{eq:D1}, it holds that 
    \begin{align}\label{eq:D-prob}
        \IP(\mathcal{D} \cap \{\|Z\|_F \leq G_0\}) \geq 1- 4\kappa.
    \end{align}  Putting these bounds back into \eqref{eq:hp-conditional}, we invoke \eqref{eq:D-prob} to conclude \eqref{eq:hp-bound-final}.

\subsection{Proof of Theorem \ref{cor:dic_behav}}
\label{pf_cor:dic_behav}
\begin{enumerate}[label=(\roman*)]
\item \emph{Low-dimensional setting.}
Recall $T=\frac{\log (N^2 L D^3)}{\log ((1-\eta_0\lambda)^{-1})}$. Note that, with $\kappa>e^{-D^2}$, we have $G_0\lesssim\mathcal{C}/\sqrt{L}$, $R\lesssim \lambda^{-1}\mathcal{C}^2\sqrt{N/L}$, $\eta_0 \asymp \frac{\lambda}{(\lambda+G^2)^2} \lesssim 1/\lambda$ and $\lambda \asymp 1 \asymp \eta_0$ from \eqref{eq:hyperparameter-choice}. Hence, the first term of \eqref{eq:low-dim} can be bounded as 
\begin{align*} 
    G_0^2(1-\eta_0\lambda)^T R^2 \lesssim \frac{\mathcal{C}^6}{NL^3D^3} \lesssim \nu^3\log^3\frac{NL}{\kappa}\frac{1}{NL^3D^3}.
\end{align*}
Moreover, from $\log(\frac{1}{\kappa})\lesssim D^2 \lesssim \log(NL)$, we have that 
\begin{align} \label{eq:firsttermfinal}
    G_0^2(1-\eta_0\lambda)^T R^2 \lesssim \nu^3(\log^3 NL)\cdot\frac{1}{NL^3}.
\end{align}
On the other hand,  write the second term  as
\begin{align}\label{eq:secondterm}
    G_0^2\sigma^2\eta_0^2D\frac{T^2}{N^2}\frac{\log(2T/\delta)}{\varepsilon^2} \lesssim \frac{\mathcal{C}^2}{L}(\mathcal{C}G+RG^2)^2D\frac{T^3}{N^2}\frac{\log(1/\delta)}{\varepsilon^2}.
\end{align}
Clearly, for $\sigma$, one obtains,
\begin{align*}
    \mathcal{C}G+RG^2 &\lesssim \frac{\mathcal{C}^2}{\sqrt{L}}\left(1 + \left( \frac{\log(NL)}{D^2} \right)^{1/2} \right) + \mathcal{C}^2\sqrt{\frac{N}{L}}\frac{\mathcal{C}^2}{L}\left( 1 + \left( \frac{\log(NL)}{D^2} \right)\right)\\
    &\lesssim \nu^2\log^3(NL)\frac{\sqrt{N}}{LD^2},
\end{align*}
where the second inequality is attained by using $\log(N/\kappa)= \log N + \log 1/\kappa \lesssim \log N + D^2 \lesssim \log NL$, and the final assertion follows from $(\log NL)/{D^2} >> (\sqrt{\log NL})/{D}$. Therefore, from \eqref{eq:secondterm}, the second term is bounded by, 
\begin{align}
    & \lesssim  \frac{\nu^5\log^7(NL)}{L}\cdot \frac{N}{L^2D^4}\cdot D\cdot \frac{\log^3(N^2LD^3)}{N^2}\cdot \frac{\log(1/\delta)}{\varepsilon^2} \nonumber \\
    & \lesssim \frac{\nu^5\log^7(NL)\log^3(N^2LD^3)}{NL^3D^3}\cdot\frac{\log{(1/\delta)}}{\varepsilon^2} \nonumber \\
    & \lesssim \frac{\nu^5\log^{10}(NL)}{NL^3}\cdot \frac{\log(1/\delta)}{\varepsilon^2}. \label{eq:secondtermfinal}
\end{align}

Combining \eqref{eq:firsttermfinal} and \eqref{eq:secondtermfinal} yields the proof for the low-dimensional case.

\item \emph{High‐dimensional setting.}
Here, $\kappa \gtrsim (NL)^{-1}$, and $D^2 \gtrsim \log(NL)$ implies that $\frac{\log(NL/\kappa)}{D^2} \lesssim 1$. We also have $G\lesssim \mathcal{C}/\sqrt{L}$, $G_0\lesssim \mathcal{C}/\sqrt{L}$ and, $R\lesssim\lambda^{-1}\mathcal{C}^2\sqrt{N/L}$. The first term of \eqref{eq:high-dim} can be bounded as 
\begin{align} \label{eq:A22}
    G_0^2R^2(1-\eta_0\lambda)^T & \lesssim \frac{\mathcal{C}^2}{L}\cdot \mathcal{C}^4\frac{N}{L}\cdot\frac{1}{\lambda^2}\cdot \frac{1}{r}  \lesssim \nu^3\log^3(NL)\cdot\frac{N}{L^2\lambda^2r}.
\end{align}
Furthermore, for the second term, observe that $\eta_0 \lesssim 1/\lambda$, and  $$\sigma= G(\mathcal{C}+RG) \asymp \frac{\mathcal{C}^2}{\sqrt{L}} + \mathcal{C}^2\sqrt{\frac{N}{L}}\cdot \frac{1}{\lambda}\cdot \frac{\mathcal{C}^2}{L}=\frac{\mathcal{C}^2}{\sqrt{L}}\big( 1+\frac{\sqrt{N}}{\lambda L}\mathcal{C}^2\big).$$
Therefore, the second term can be bounded as 
\begin{align}
    G_0^2\sigma^2\eta_0^2D\frac{T^3}{N^2}  \leq \frac{\mathcal{C}^2}{L}\cdot \sigma^2\frac{1}{\lambda}D\cdot \frac{T^3}{N^2} & \leq \frac{\mathcal{C}^2}{L}\cdot \log^3{r}\cdot \frac{D}{N^2\lambda^2}\cdot \sigma^2 \nonumber\\
    & \leq \frac{\mathcal{C}^2}{L}\log^3r\frac{D}{N^2\lambda^2}\cdot\frac{\mathcal{C}^4}{L}\biggl( 1 + \frac{N}{\lambda^2L^2}\mathcal{C}^4 \biggr) \nonumber\\
    & \leq \frac{\mathcal{C}^6D}{N^2L^2\lambda^2}\biggl( 1 + \frac{N}{\lambda^2L^2}\log^2(NL) \biggr) \log^3r \nonumber\\
    & \lesssim \nu^3 \frac{N\log^3 r}{L^2\lambda^2}\log^3 NL \frac{D}{N^3}\biggl( 1 + \frac{N}{\lambda^2L^2}\log^2NL \biggr) . \label{eq:A23}
\end{align}
Assertions \eqref{eq:A22} and \eqref{eq:A23} conclude the proof.

 \end{enumerate}

\subsection{Proof of Proposition \ref{prop1}} \label{pf_prop1}
    From $\lambda \asymp N/D\asymp \sqrt{N}$, it follows that $\frac{D^2}{NL^2}\asymp \frac{N}{L^2\lambda^2}\asymp N^{-1}$, and hence, $\frac{\log^2 (NL)}{\lambda^2} \lesssim 1$. Therefore, from \eqref{eq:high-dim}, \eqref{eq:over-parametrize} follows trivially. Further, the first term in \eqref{eq:over-parametrize} dominates as long as $r \log^3 r \ll \frac{N^3}{D}$, yielding \eqref{eq:early-stopping} when $r\asymp N$ .

\subsection{Proof of Theorem \ref{thm:robustness-thm}}
\label{pf_thm:robustness-thm}
    Recall the definition of $G_0$ and $R$ from Theorem \ref{thm:main_thm}. In view of $\kappa>N\exp(-D^2)$, \eqref{eq:bound-on-unprojected-Z} and $\| \hat{\Gamma} \|_F \vee \| \Gamma^{T^\bad} \|_F \leq R$, using \eqref{eq:bound_indv_z}, it holds that 
    \begin{align}\label{eq:truncation-effect}
       (\langle \hat{\Gamma} , Z\rangle  - \langle \hat{\Gamma}_\bad, Z\rangle)^2 \leq \frac{\mathcal{C}^2}{L} R^2 ,
    \end{align}
    with probability at least $1-\kappa$. On the other hand, for the analysis of the ridge estimates, recall \eqref{eq:ridge}. Clearly, from Lemma \ref{lemma:R-term}, it holds with probability $\geq 1-2\kappa$ that
    \begin{align}\label{eq:bound-unperturbed}
        \left\|\sum_{k=1}^N \operatorname{vec}(Z_k) \operatorname{vec}(Z_k)^\top\right\|_F \leq {\mathcal{C}^2}\frac{N}{L} < N\lambda,
    \end{align}
    where the final equality follows from $\lambda> \mathcal{C}^2L^{-1}$. Moreover, from Lemma \ref{lemma:G-term}, it holds with probability $\geq 1-2\kappa$ that 
    \begin{align}\label{eq:bound-perturbed}
        \left\|\operatorname{vec}(Z_{\bad, i}) \operatorname{vec}(Z_{\bad, i})^\top\right\|_F 
 \lesssim \frac{\mathcal C^2}{L}+\frac{\alpha^2 \mu^4}{L}\asymp \frac{\alpha^2 \mu^4}{L} \lesssim N\lambda,
    \end{align}
    where the first part of the inequality follows from the lower bound on $\alpha^2\mu^2$ and the second inequality follows from the upper bound on $\alpha^2\mu^4$ as stated in \eqref{eq:lowerbound}. Consequently, combining \eqref{eq:ridge} with \eqref{eq:bound-unperturbed} and \eqref{eq:bound-perturbed} jointly yields,
    \begin{align}\label{eq: intermediate-ridge}
        \|\operatorname{vec}(\Gamma^\star_\bad - \Gamma^\star) \|\geq \frac{\| y_{i, L+1}' \operatorname{Vec}(Z_i') - y_{i, L+1} \operatorname{Vec}(Z_i)\|}{N\lambda}
    \end{align}
    with probability at least $1-4\kappa$. Since $\|y_{i,L+1} \operatorname{vec}(Z_i)\|\leq \frac{\mathcal{C}^2}{\sqrt{L}}$ with probability at least $1-\kappa$, invoking \eqref{eq:lowerbound}, yet another application of Lemma \ref{lemma:G-term} yields
    \begin{align} \label{eq:numerator-lowerbound}
        \| y_{i, L+1}' \operatorname{Vec}(Z_{\bad, i}) - y_{i, L+1} \operatorname{Vec}(Z_i)\| \geq \alpha^2\mu^2
    \end{align}
    with probability at least $1-\kappa$. Since \eqref{eq:lowerbound} also implies $\frac{\alpha^2\mu^2}{N\lambda}> R^2\frac{\mathcal{C}^2}{L}$, from \eqref{eq:truncation-effect}, \eqref{eq:bound-perturbed}, \eqref{eq: intermediate-ridge}, and \eqref{eq:numerator-lowerbound}, we obtain \eqref{eq:robustness-bound}.

\subsection{Auxiliary Lemmas}
The following lemmas are instrumental to proving our theorems \ref{thm:main_thm} and \ref{thm:robustness-thm}, and hereby are listed. In particular, Lemma \ref{lemma:C-term} and \ref{lemma:R-term} follows using Hoeffding's inequality and a union bound argument.
\begin{lemma}\label{lemma:C-term}
    If $z_{kj}\sim N(0,1+\tau^2)$ $k\in [N], j\in [L]$ are not necessarily independent, then 
    \[ \IP\bigg(\max_{k,j} |z_{ij}| \lesssim \sqrt{(1+\tau^2)\log \left(\frac{4NL}{\kappa}\right) } \bigg) \geq 1- \kappa. \]
\end{lemma}

For the Lemmas \ref{lemma:G-term} and \ref{lemma:R-term}, note that for any vector $x \in \R^D$, the Euclidean norm $\|x\|_2=\sup_{a \in \mathbb{S}^{D-1}}a^\top x$. For any fixed $a \in \mathbb{S}^{D-1}$ and $k \in [N]$, $a^\top \sum_{i=1}^L x_{k,i}$ is a sub-Gaussian random variable with variance proxy $L/D$ . Therefore
\[
\mathbb P\left[\bigg|a^\top \sum_{i=1}^L x_{k,i}\bigg|>\sqrt{L/D}\,t\right] \lesssim \exp\left(-t^2\right).
\]

Therefore, using a covering number argument similar to Theorem 1.19 of \cite{rigollet2023high} one can show the following.

\begin{lemma}\label{lemma:G-term}
    Suppose $(x_{k,i})_{k \in [N], i \in [L]} \overset{i.i.d.}{\sim} \mathcal{U}(\mathbb{S}^{D-1})$. Then,
    \[ \IP\bigg(\max_{k \in [N]} \|\sum_{i=1}^L x_{k,i}\|_2 \lesssim \sqrt{L}(1 + D^{-1/2} (\log (\frac{N}{\kappa}))^{1/2})\bigg) \geq 1 - \kappa. \]
\end{lemma}

\begin{lemma}\label{lemma:R-term}
    Suppose $(x_{k,i})_{k \in [N], i \in [L]} \overset{i.i.d.}{\sim} \mathcal{U}(\mathbb{S}^{D-1})$. Then,
    \[\IP\bigg(\|\sum_{k=1}^N \sum_{i=1}^L \operatorname{vec}(x_{k, L+1}x_{k,i}^\top)\|_2 \lesssim \sqrt{NL}(1 + D^{-1} (\log(\frac{1}{\kappa}))^{1/2})\bigg) \geq 1- \kappa. \]
\end{lemma}

\section{Numerical Experiments Details}
This section details and extends upon the numerical examples presented in Section \ref{se:simulation}.
\subsection{Effect of privacy on prediction risk: low- vs. high-dimensional regimes} \label{se:N vs eps_appn}
In Section \ref{se:N vs eps}, we empirically investigate how the level of privacy, parameterized by $\varepsilon$, affects the prediction accuracy of \fancyname\ through its impact on the excess risk.
\paragraph{Low-dimensional regime.}We first consider the low-dimensional regime with feature dimension fixed at $D = 5$. Training set sizes are varied over $N \in \{1000, 1500, 2000, 2500, 3000, 3500, 4000\}$, with prompt length set as $L = \lfloor \sqrt{N} \rfloor$ and privacy levels $\varepsilon \in \{0.2, 0.4, 0.6, 0.8, 1.0\}$. The hyperparameters $\mathcal{C}, \mathcal{G}, \mathcal{R}$ are chosen according to Theorem~\ref{thm:main_thm}, with $\kappa = 1$ and $\delta = 10^{-5}$. The step size is set as $\eta_0 = 3.17 / (5 + G^2)^2$, where $G$ denotes an upper bound on the norm of the projected features $\wt Z$, and the ridge regularization parameter is fixed at $\lambda = 5$. We work in a noiseless setting with $\tau^2 = 0$.

For each $(N, \varepsilon)$ pair, we generate $N$ prompts according to \eqref{eq:def_prompt} and run $T$ iterations of \fancyname, where $T = \log N^{5/2} / \log(1 - \lambda \eta_0)$, as prescribed by Theorem~\ref{cor:dic_behav}. Test performance is evaluated on $n_{\operatorname{test}} = 500$ held-out prompts. Each test prediction is computed as $\langle \hat{\Gamma}, Z_{\operatorname{test}} \rangle$, where $Z_{\operatorname{test}}$ is constructed using \eqref{eq:construct_meta_varb}. The excess risk is defined as
\begin{align}
\label{eq:avg_test_err}
\mathcal{E}_{\operatorname{test}} = \frac{1}{n_{\operatorname{test}}} \sum_{k=1}^{n_{\operatorname{test}}} \big(\langle \hat{\Gamma} - \Gamma^\star, Z_{k,\operatorname{test}} \rangle \big)^2,
\end{align}
where $\Gamma^\star$ denotes the non-private ridge estimator trained on the same data. We repeat the entire procedure $B = 500$ times and report the average excess risk $\bar{\mathcal{E}}_{N, \varepsilon}$.

\begin{figure}[t!]
        \centering
        \includegraphics[scale=0.4]{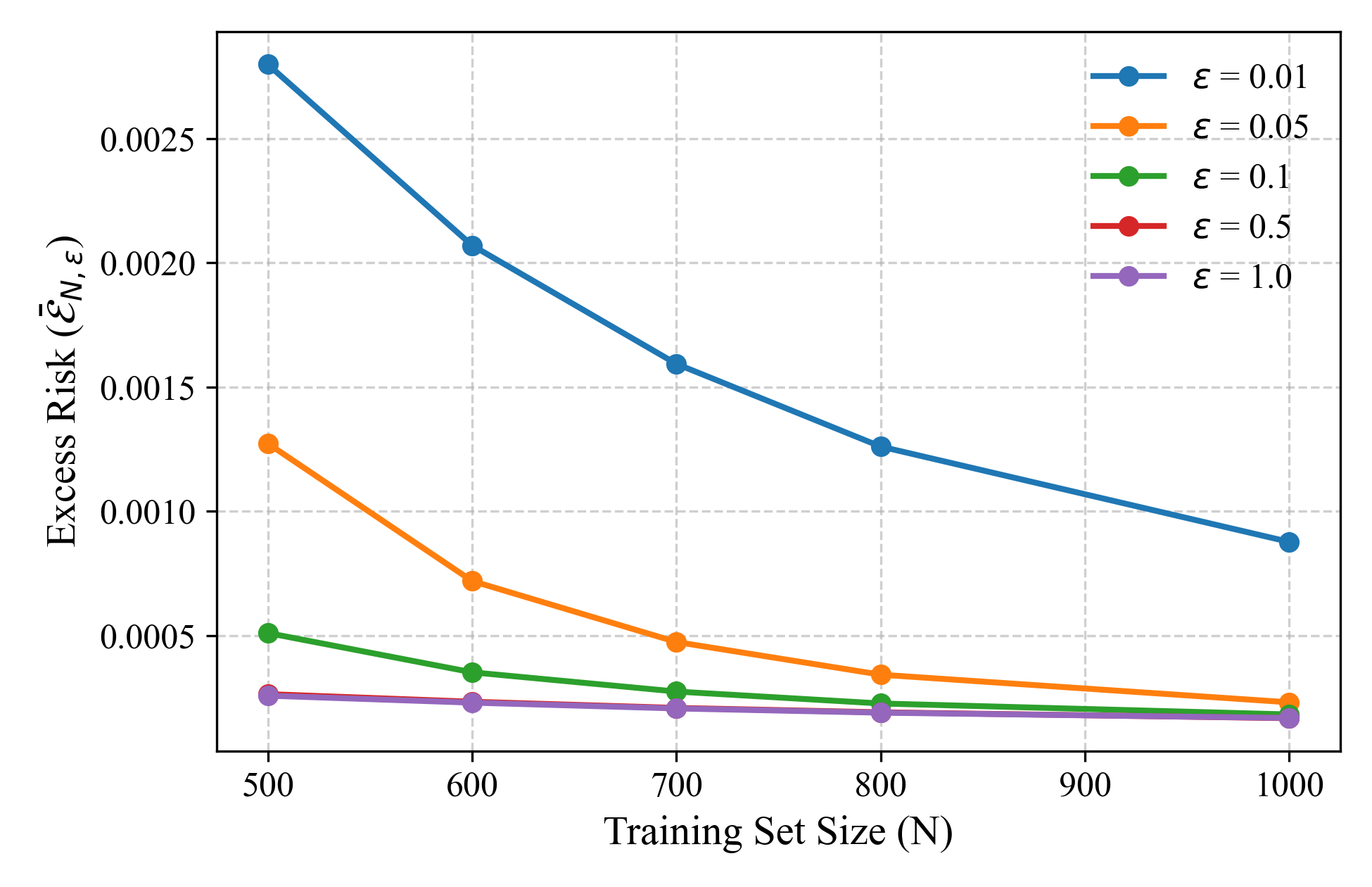}
        \label{fig:high_dim_vs_N_eps_right}
    \caption{Excess risk of \fancyname\, as a function of training set size $N$ for different values of the privacy parameter $\varepsilon$ with $D=\lfloor\sqrt{N}\rfloor$. }
    \label{N_eps_combined}
\end{figure}

The left panel of Figure~\ref{N_eps_combined} illustrates that for each fixed $\varepsilon$, the excess risk decreases with $N$ at a super-quadratic rate, in agreement with Theorem~\ref{cor:dic_behav}(i). For fixed $N$, the excess risk also decreases with increasing $\varepsilon$, highlighting the trade-off between privacy and predictive accuracy in this regime.

\paragraph{High-dimensional regime.}
We also consider the high-dimensional regime where $D \asymp L \asymp \sqrt{N}$. We vary $N \in \{500, 600, 700, 800, 900, 1000\}$ and $\varepsilon \in \{0.01, 0.05, 0.1, 0.5, 1.0\}$. The ridge regularization parameter is set as $\lambda = N/D$, and we use a fixed number of iterations $T = 5$ with step size $\eta = 0.07 ND / (N + DG^2)^2$. All other parameters mirror those used in the low-dimensional setting. The average excess risk, computed over $B = 500$ repetitions, is reported in the right panel of Figure~\ref{N_eps_combined}. The excess risk decreases with both $N$ and $\varepsilon$, though at a slower rate than in the low-dimensional case, consistent with Theorem~\ref{cor:dic_behav} and reflecting the increased challenge of private learning in high dimensions.

\subsection{Effect of early stopping in over-parametrized setting} \label{se:over-parametrized_appn}
In Section \ref{se:over-parametrized}, we investigate how the number of gradient descent steps $T$ affects the test performance of the linear attention head trained using \fancyname. We fix $N=1000$ and consider the overparameterized setting with $L \asymp D \asymp \lfloor \sqrt{N} \rfloor$, under fixed privacy parameters $\varepsilon = 0.8$ and $\delta = 10^{-5}$. The step size is set as $\eta_0 = 0.007 \lambda / (\lambda + G^2)^2$, and remaining hyperparameters follow the setup from the previous experiment. We vary $T$ over $\{1, 20, 40, \ldots, 480\}$ and compute the average test error over $500$ held-out prompts, repeated over $B=500$ independent trials.

Figure~\ref{fig:phase-transition} plots the evolution of two components of the prediction error: the \emph{cost of descent} (blue) incurred by underoptimization, and the \emph{cost of privacy} (orange) due to noise injection. For small $T$, the descent cost dominates and the error decreases with additional optimization. However, beyond a critical number of iterations, the cost of privacy dominates, causing error to increase as more noise accumulates. This trade-off, predicted theoretically in Remark~\ref{rem:opt_stop}, yields a phase transition in the test error under privacy constraints. In contrast, in the noiseless setting (approximating ridge regression), the error decreases monotonically with $T$.

\subsection{Robustness of \fancyname} \label{se:robustness-simu_appn}

In Section \ref{se:robustness-simu} we evaluate the robustness of \fancyname\ under adversarial perturbations, following the setup in Section~\ref{se:robustness}. A single training prompt is perturbed by adding $1$ to all features and $\alpha = cN^p$ to all responses, with $c \in \{2, 4\}$ and $p \in \{2, 2.02, 2.04, 2.06, 2.08, 2.1\}$. We fix $N = 5000$, $L = 500$, $D = 5$, $\varepsilon = 0.5$, and $\delta = 10^{-2}$. Generalization error is measured via \eqref{eq:robustness-bound}. We compare the ridge estimator $\Gamma^\star$ with the output of \fancyname\ after $T = \log N$ iterations, using $\lambda = 0.01$ and step size $\eta_0 = 0.007 / (0.01 + G^2)^2$, with all other parameters unchanged. Figure~\ref{fig:risks_comparison} reports the average prediction error over $500$ test prompts, averaged over $500$ trials. As $p$ increases, ridge regression becomes increasingly sensitive to the perturbation, while differentially private pretraining with \fancyname\ remains substantially more robust.

\end{document}